# Sequence-Agnostic Multi-Object Navigation

Nandiraju Gireesh[*,1], Ayush Agrawal[*,1], Ahana Datta[*,1], Snehasis Banerjee[1,2], Mohan Sridharan[3]
Brojeshwar Bhowmick[2], Madhava Krishna[1]

[1]Robotics Research Center, IIIT Hyderabad, India
[2]TCS Research, Tata Consultancy Services, India
[3]Intelligent Robotics Lab, University of Birmingham, UK

*Abstract*— The *Multi-Object Navigation* (MultiON) task requires a robot to localize an instance (each) of multiple object classes. It is a fundamental task for an assistive robot in a home or a factory. Existing methods for MultiON have viewed this as a direct extension of *Object Navigation* (ON), the task of localising an instance of one object class, and are *pre-sequenced*, i.e., the sequence in which the object classes are to be explored is provided in advance. This is a strong limitation in practical applications characterized by dynamic changes. This paper describes a deep reinforcement learning framework for *sequence-agnostic* MultiON based on an actor-critic architecture and a suitable reward specification. Our framework leverages past experiences and seeks to reward progress toward individual as well as multiple target object classes. We use photo-realistic scenes from the Gibson benchmark dataset in the AI Habitat 3D simulation environment to experimentally show that our method performs better than a *pre-sequenced* approach and a state of the art ON method extended to MultiON.

## I. INTRODUCTION

Consider the home environment in Figure 1 with instances of object classes such as *bed* and *toilet* in different rooms. A core task for an assistive robot in such an environment is to locate instances of specific object classes. This *Multi-Object Navigation* (MultiON) task [2] is a generalization of the *Object Goal Navigation* (ON) task [3], [4] that requires a robot to find an instance of a single object class. Such ON and MultiON tasks arise in many practical applications.

Humans perform MultiON tasks with seemingly little effort. For example, a human going for a walk may need a pair of socks, their house keys, and an umbrella. To locate these objects, they build on prior experience in this and other related (home) environments, to explore a series of locations likely to contain one or more of these objects. Also, humans try to concurrently minimize the distance to all target object classes and adapt their exploration based on observations. For example, if keys are observed unexpectedly on top of the cabinet while searching for socks, the human will stop and confirm whether these are the house keys. State of the art methods for MultiON, on the other hand, focus on ON tasks or perform **P**re-**S**equenced **M**ultiON (PSM) in which the robot is given the sequence in which the target object classes are to be explored [2]. For example, the robot in Figure 1

*Denotes equal contribution

(left sub-figure) is given the sequence $\{chair, toilet, couch\}$ as the MultiON task. It first searches for and finds a chair; although it spots a couch as it moves near the chair to confirm the chair's location. It then searches for a toilet (second object class in the sequence) before coming back to confirm the location of the couch seen earlier.

Inspired by insights from human cognition, we present a framework for **S**equence **A**gnostic **M**ultiON (SAM); the robot is neither provided nor forced to compute a global order in which it locates instances of the target object classes. Instead, the robot explores likely locations of the target objects and automatically adapts its exploration based on observations. In Figure 1 (right sub-figure), for example, the robot observes a chair and a couch nearby. It first confirms the location of the couch before confirming the chair's location, then explores further to locate a toilet; the distance traveled and the task completion time are substantially less compared with PSM in Figure 1 (left sub-figure). Specifically, our framework makes two key contributions:

1) Instead of computing the globally optimal sequence of trajectories by evaluating all possible paths through locations in the domain, the robot builds on past experience in environments with a similar distribution of regions and objects, greedily choosing a series of 'long-term goals' in an attempt to concurrently minimize the distance to an instance of all target object classes.

2) Extends our previous work on ON [5] to develop a deep reinforcement learning (RL) framework for SAM, introducing a novel reward specification and adapting the actor-critic network to reward the concurrent progress to instances of multiple object classes, instead of only rewarding the progress towards identifying an instance of a single object class.

We experimentally evaluated our framework through ablation studies, and quantitative and qualitative comparisons with relevant baselines using photo-realistic scenes from the Gibson benchmark dataset in AI Habitat, a 3D simulation environment [6]. We computed five standard performance measures over experimental trials involving different number of object classes. These experiments demonstrated the significantly better performance provided by our framework, compared with the use of a predetermined sequence of object classes, and with two other methods for selecting long-term

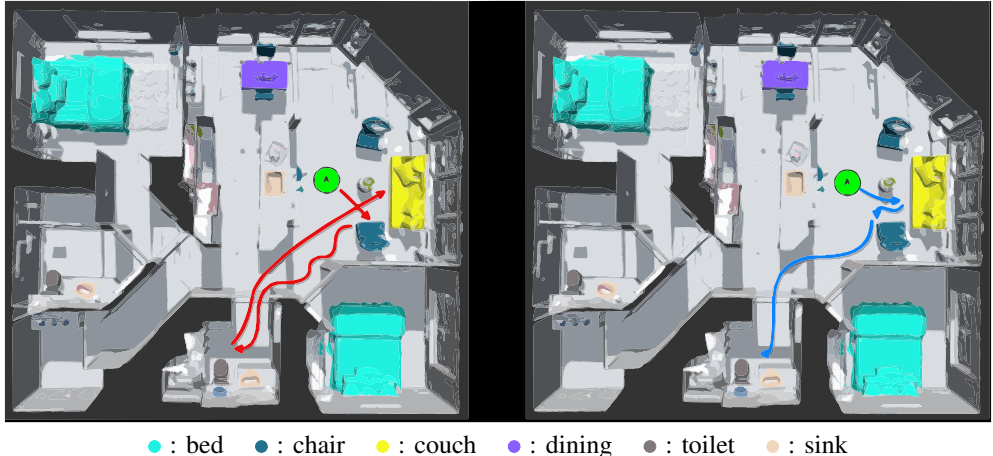

● : bed   ● : chair   ● : couch   ● : dining   ● : toilet   ● : sink goals: random search, and a state of the art deep RL method for ON [4] extended to MultiON. In particular, our SAM framework provided ≈ 50% reduction in the number of time steps and path length compared with a PSM baseline.

Fig. 1: Example trajectories of the same episode when agent traverses in Pre-Sequenced MultiON (PSM) and Sequence Agnostic MultiON (SAM). In this particular episode, the goal objects are specified as {chair, toilet, couch}. Paths taken by PSM and SAM are shown in red and blue. Semantic annotations for the Gibson Tiny split [1] has been used here.

## II. RELATED WORK

We review related work in Embodied AI tasks, single object search tasks, and multi-object search tasks.

**Embodied AI tasks.** Research in Embdodied AI has picked up pace as data sets containing 3D scenes of indoor environments have become readily available, e.g., Matterport3D [7], Gibson [8], and Habitat-Matterport 3D [9]. These datasets get loaded in AI simulators like Habitat [6] and GibsonEnv [8], and support the exploration of problems such as: (a) PointGoal Navigation [10], [11], which requires a robot to move to a specified point location in the domain; (b) ObjectGoal Navigation [4], [5], [12], [13], which requires a robot to find a target object instance in an unexplored environment; and (c) Visual Language Navigation [14], which requires a robot to navigate to target scene based on complex instructions and scene descriptions.

**Object search tasks.** The aim of object search tasks (e.g., Object Navigation, ON) is to navigate to an instance of a specific target object class in a previously unseen environment as quickly as possible. End-to-end reinforcement learning methods are the state of the art for mapping pixels directly to actions suitable for accomplishing this task [3], [15]. These methods find it difficult to generalize to previously unseen scenarios since they do not build a semantic representation of the environment. In contrast, methods that seek to promote better generalization construct an allocentric map that encodes semantic priors [4], [16]–[18].

**Multiple object search tasks.** The goal of the MultiON task [2] is for a robot to localize an instance (each) of more than one object class. The original challenge description introduced colored cylinders in the Habitat environment, with only one instance of each object class, and provided the sequence in which instances of the target object classes were to be localized. This reduced MultiON to a fixed sequence of ON tasks, which we refer to as pre-sequenced MultiON (PSM). Methods for PSM have predominantly developed and evaluated different map representations and memory architectures. For example, the original paper on MultiON explored *NoMap*, *Oracle Map*, and *Learned Map* agents under the assumption of noiseless pose estimation [2]. Also, their RL policy rewarded the agent's progress toward, as well the localization of, an instance of the target object class under consideration. Another paper sought to decouple mapping from localization, with the policy being a combination of exploration and moving towards a particular object class instance [19]. There has also been work exploring the role of auxiliary tasks in improving the navigation performance, specifically by taking into account object instances that had already been observed while executing the policy [20].

In contrast to existing work, we allow more than one instance of each object class, consider naturally-occurring object classes in the Gibson indoor scenes instead of artificially-induced objects. We also relax the strong limitation imposed by the PSM formulation, and instead focus on Sequence-Agnostic MultiON. We do so by adapting our prior deep network architecture for ON [5] to consider embeddings of additional inputs and a novel reward specification.

## III. PROBLEM FORMULATION AND FRAMEWORK

This section describes the Sequence-Agnostic MultiON (SAM) task and our proposed framework.

### A. Task Description

In each episode of SAM, the robot must locate an instance (each) of a set $G$ of one or more target object classes in the environment. The environment consists of at least one (and often two or more) instance(s) of each object class $G_i$. At each timestep $t$ in the episode, the robot receives: (a) egocentric RGB and depth observations of the scene within the robot's view; (b) the robot's pose in the domain; and (c) an one-hot encoding for each object class (out of $N = 16$ classes) whose instance is to be located. We use $k$-ON to refer to an episode with $k$ target object classes. The robot does not have any prior map of the domain or the sequence in which the target object classes are to be explored; if such a sequence is provided, it is a PSM task. An object class is considered to be $found$ when the robot navigates close (distance to success or target, $d_s \leq 1m$) to an instance of the class; if no such instance is found within a maximum number of timesteps, the episode is said to be unsuccessful. The robot can execute one of these four actions: {*move-forward, turn-left, turn-right, stop*}. The *move-forward* action moves the robot forward by $0.25m$, whereas the *turn* actions cause the robot to rotate by $30°$ in the appropriate direction (*left, right*).

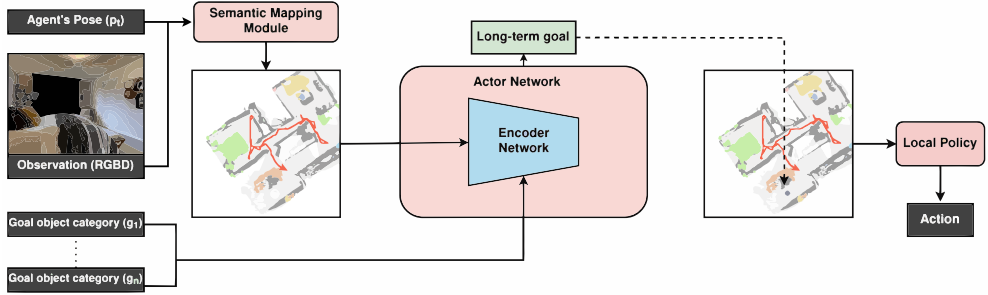

Fig. 2: Our framework consists of three main components. The *Semantic Mapping* module uses odometry pose readings and RGB-D observations to create an allocentric semantic map of the local environment. The *Encoder Network* extracts high-level feature embeddings from the semantic map and encoding of target classes. These features are used to train an actor-critic network that outputs a long-term goal to search and find an instance of the target object classes. Analytical planners used by the deterministic *Local Policy* compute low-level navigation actions to reach a given long-term goal.

### B. Proposed Framework

As stated earlier, we pursue a deep RL formulation of SAM and present a modular approach based on the actor-critic architecture. Figure 2 provides an overview of this framework, which extends our prior work on ON [5], and comprises three modules:

1) **Semantic Mapping:** builds a map of the domain (i.e., metric arrangement of space with obstacles and empty space) from the RGB-D (RGB and depth) data and pose observations. Robot uses the map to localize itself, and processes the RGB-D data to identify and localize specific object instances in this map.
2) **Encoder Network:** Receives as input the semantic map (above) and the encodings of target object classes, and extracts high-level features. These feature are sent to an Actor-Critic network that repeatedly computes a 'long-term goal', i.e., a region the robot should travel to in search of an instance of a target object class.
3) **Deterministic Local Policy:** Uses analytical planners to compute the low-level navigational actions that need to be taken to reach the current long-term goal region.

The semantic map constructed by the Semantic Mapping module is a matrix of dimension $K \times M \times M$. This contains $K$ channels of $M \times M$ size maps where $K = C + 2$ and $C$ is total number of semantic categories. The first two channels contain the obstacles and the explored areas respectively while the remaining channels contain the $C$ object categories. We use the mapping procedure from a state-of-the-art method for ON [4]. A pretrained Mask R-CNN model [21] is used to estimate the semantic categories from the observed RGB information. The depth observations are used to compute point clouds and each point in the point cloud is associated with the estimated semantic categories. With the help of differentiable geometric computations over each point in the point cloud, we build a voxel representation that is then converted into a semantic map of dimension $(C+2) \times M \times M$. We also use random shift augmentation on the predicted semantic map to promote generalization.

The encoder network takes as input the estimated semantic map from the previous module, the robot's current and past locations, the objects found and localized so far, and the encoding of the target object classes. *The one-hot encoding of multiple object classes was not considered in our prior work* [5]. High-level feature embeddings are extracted by the encoder which are then used by the actor network to obtain a long-term goal, i.e., the next location the robot should move for the target object search. The encoder network comprises 4 convolutional layers with $3 \times 3$ kernels and 32 channels [22]. ReLU activation is applied after each convolutional layer. A stride length of 1 is used everywhere. The output of these layers is passed to a fully-connected layer normalized by the LayerNorm operation [23]. Also, a hyperbolic tangent non-linear transform is applied to the 50-dim output of the fully-connected layer. The weight matrix of the fully connected layer and the convolutional layers is initialized by orthogonal initialization [24] with bias set as zero.

The output of the encoder network is used by the actor-critic network. The actor and critic components work with the same weights in the convolutional layers but have separate encoders. The weights in the convolutional layers are allowed to be updated only by the optimizer in the critic network. We employ the clipped double Q-learning method [25] for the critic. In this method, each Q-function is parameterized as a three-layer multi-layer perceptron (MLP) with ReLU activations after each layer except the last one. All the transition states are stored using a replay buffer. The transition states include the semantic map, target object classes, action, reward, next semantic map, and the subsequent target object classes. A set of transition states are obtained from the replay buffer and, along with the augmented semantic map, given as input to the encoder and actor-critic networks. Every 25 timesteps, a new long-term goal is sampled. Note that both the actor and critic networks' parameters are revised during training. Once trained, only the actor network is used for testing. For ease of understanding, only the actor network is labeled in Figure 2. The specification of the reward function, a key contribution of this paper, is described in Section III-C.

When a long-term goal is provided by the actor-critic network, the local policy module uses the Fast Marching Method [26] to guide the robot to this region. Specifically, the obstacle channel from the semantic map generated in the semantic mapping module is used to compute the shortest path from the current location to the current long term goal. The robot then computes the low-level navigational actions to navigate along the computed shortest path.

### C. Reward Function

A key contribution of this work is the reward specification used to train our deep network. Recall that our objective is to mimic the intuitively appealing sequence-agnostic behavior of humans engaged in MultiON tasks. In order to do so, we identified three desired characteristics that we wanted our reward function to capture:

1) We wanted to encourage the robot to find an instance of each target object class.
2) We wanted to motivate the robot to concurrently reduce the distance to an instance of more than one object class. However, we did not want it compute the globally optimal exploration sequence by considering all possible sequences because that could be computationally intractable in practical deployment.
3) We wanted to encode *non-procrastination*, i.e., minimize the time spent looking for object instances.

Based on these desired characteristics, we formulated the reward function as follows:

$$Reward = R_{\text{sub-goal}} + \alpha_{process} * R_{process} + CNR \quad (1)$$

where $R_{sub-goal}$ is the reward for achieving a 'sub-goal', i.e., an instance of one of the target object classes; $R_{process}$ is the process reward; and CNR is the negative reward, i.e., cost (currently $-0.01$) accumulated at every timestep. The value of CNR was set such that the penalty accrued over a typical episode was small relative to the other parts of the reward function. We used scaling factor $\alpha_{process} = 0.1$ (determined experimentally) to vary the relative influence of $R_{process}$ on the overall behavior of the robot.

*1) Sub-goal reward:* $R_{\text{sub-goal}}$ is the standard reward the robot receives when it localizes an instance of any target object class. To ensure that this reward is only received at the corresponding timestep, we modeled it as:

$$R_{sub-goal} = 1_{sub-goal} * r_{\text{sub-goal}} \quad (2)$$

where $1_{sub-goal}$ is an indicator function that is equal to 1 *iff* the robot reaches an instance of one of the target object classes, and $r_{sub-goal}$ is the instantaneous real-valued reward. This can be restated as:

$$R_{sub-goal} = \begin{Bmatrix} r_{sub-goal} & \text{if a sub-goal is reached} \\ 0 & \text{otherwise} \end{Bmatrix} \quad (3)$$

We experimentally set $r_{sub-goal} = 2$ to be relatively higher than the other two parts of our reward, in order to enable the robot to reach the sub-goal with higher priority.

*2) Process reward:* We recognized that this part of the reward function may require a trade-off with the first part of the reward ($R_{\text{sub-goal}}$), e.g., focusing on a shortest path to a particular region may help the robot obtain $r_{\text{sub-goal}}$ as soon as possible but it may make sense to deviate from this region to another region nearby if an instance of another target object class is likely to be found there.

We first used the known (i.e., ground truth) location of object instances during training to compute the distance to the closest instance of each target object class at each timestep. We used this information to compute, at each timestep $t$, the total decrease in the geodesic distance to the nearest instance of each target object class $g_i$:

$$d_t = \sum_{i}^{N}(dtg_{i,t-1} - dtg_{i,t}) \quad (4)$$

where $dtg_{i,t}$ refers to the shortest distance to an instance of object class $i$ at timestep $t$; and $N$ is the number of target object classes whose instance remains to be localized in this episode. Once $d_t$ is computed, $R_{process}$ is computed as:

$$R_{process} = \begin{Bmatrix} \frac{n}{N} + d_t & \text{if } dtg \text{ of } n \text{ classes decreases} \\ d_t & \text{otherwise} \end{Bmatrix} \quad (5)$$

where the additional reward received depends on the fraction of the target object classes to whose instances the robot was able to reduce its distance during the training episode. This part of the reward thus encourages the robot to greedily attempt to concurrently localize more than one object based on prior experience in similar environments and on the observations received in the current episode. Note that *the reward function's components remain the same irrespective of the number of target object classes*. Experimental results (below) demonstrate the benefits of this reward specification and our SAM framework.

### IV. EXPERIMENTAL SETUP

We evaluated our SAM framework's capabilities using photo-realistic benchmark scenes (3D reconstructions of home environments) from the Gibson dataset in the AI Habitat simulator [6]. We used the standard ObjectNav Challenge's 25 scenes during training by setting up $\approx 1000$ episodes (total) of $k$-ON task, randomly selecting $k \in [2, 3]$ target object classes in each episode. The robot's starting position is randomly sampled from navigable points in the environment in each episode. For testing, we generated datasets for 2-ON and 3-ON tasks using five scenes from the Gibson dataset that were not considered during training. This testing dataset included 200 episodes for each of the five scenes, resulting in a total of 1000 episodes. We did not explore k-ON tasks for $k > 3$ in this benchmark dataset, as the number of scenes containing one or more instances of the target object classes were less for higher $k$.

As stated in Section III-A, the robot's observations were in the form of $4 \times 640 \times 480$ RGB-D images, and the success threshold $d_s = 1m$. The maximum episode length was 1000 steps and 600 for the 3-ON and 2-ON episodes respectively. The target object classes were those considered by the state of the art approaches for ON [4]: 'chair', 'couch', 'potted plant', 'bed', 'toilet' and 'tv'. We experimentally evaluated the following hypotheses about our framework:

**H1:** *Our SAM framework traverses shorter paths and takes fewer timesteps when compared with PSM.*

**H2:** *Our SAM framework provides better long-term goals than the state of the art object navigation baseline extended to multi-object settings.*

We considered three baselines for evaluation:

1) **Random:** The robot pursued the SAM task (i.e., no fixed object class sequence) but it chose an action randomly in each timestep of each episode. This is a standard comparative benchmark in literature to understand the margin of improvement by other methods.
2) **PSM:** The robot was given an order (i.e., sequence) in which it had to explore the target object classes; the underlying framework was the deep RL approach we used for ON in our prior work [5].
3) **Multi-Semantic Exploration (M-SemExp):** The robot pursued the SAM task but the underlying deep RL method extended a state of the art method developed for ON [4] to the MultiON setting. In particular, reward was specified as the total decrease in geodesic distance to the nearest instance of each object $g_i$.

$$R_{SemExp} = \alpha_{SemExp} * \sum_{i}^{N}(dtg_{i,t-1} - dtg_{i,t}) \quad (6)$$

where $dtg_{i,t}$ is the shortest distance to an instance of object class $g_i$ at timestep $t$; $N$ is the number of target object classes whose instance remains to be localized.

We have considered five standard performance measures:
(i) **Success (%):** Fraction of episodes in which the robot successfully localizes an instance of each target object class within the maximum number of steps allowed.
(ii) **Sub-success (%):** Ratio of the number of target object classes whose instance has been localized to the total number of target object classes in an episode.
(iii) **Timesteps:** The number of timesteps taken to successfully complete a particular episode. A forward move or a rotation move counts as a timestep.
(iv) **Global Path Length (m):** The length of the path (in meters) traversed by the robot to successfully identify an instance of each target object class in an episode.
(v) **Global-SPL (G-SPL, %):**

Ratio of the globally minimum path length ($g$) and the length of the actual path taken by our robot in a particular episode; $g$ is computed as the shortest path that visits one instance of each target object class.

| Scene Name | Timesteps ↓ | | Global Path Length (m) ↓ | |
|---|---|---|---|---|
| | PSM | SAM | PSM | SAM |
| Collierville | 242 | **122** | 29.53 | **16.16** |
| Corozal | 336 | **179** | 46.23 | **27.28** |
| Darden | 248 | **117** | 31.43 | **16.08** |
| Markleeville | 272 | **140** | 35.41 | **18.87** |
| Wiconisco | 389 | **224** | 52.81 | **33.94** |

TABLE I: Our SAM framework provides significantly better performance than the PSM framework; with numbers averaged over 200 paired episodes for each of the five scenes.

$$\text{G-SPL} = (success) * \frac{g}{max(g, p)} \quad (7)$$

where $p$ is the length of the actual path traversed by the robot to localize an instance of each target object class. Furthermore, we included some qualitative results (see below).

## V. EXPERIMENTAL RESULTS

This section describes the qualitative and quantitative results of the experimental evaluation of our SAM framework.

**Qualitative Results:** Figure 3 shows a qualitative comparison of our SAM framework with the PSM baseline in the form of snapshots for a specific episode. Recall that PSM is provided the sequence {couch, tv, toilet} in which the target object classes are to be explored whereas only the target object classes are known in the SAM formulation. With our SAM framework, the robot quickly moves toward and localizes a toilet in timestep 21, a couch by timestep 72, and a TV by timestep 81. With PSM, on the other hand, the robot takes more than 200 timesteps to complete this task. These results partially support **H1**.

**Quantitative Results:** We then evaluated **H1** quantitatively, comparing our SAM framework with the PSM baseline, with the results averaged over the successful episodes in 200 *paired* episodes of five scenes summarized in Table I. Note that both frameworks had the same environment and the same robot starting position in each paired episode. To facilitate a fair comparison, we first ran each episode with SAM, i.e., with no constraint on the order in which an instance of the target object classes is to be found. The order in which the robot ended up localizing the target object classes was then provided as the target sequence for the PSM approach. As shown in Table I, our SAM framework provided an $\approx 50\%$ reduction in the average number of timesteps and the average path length compared with the PSM framework. This improvement in performance with our SAM framework was strongly influenced by our design of the reward function for the deep network architecture; see Section III-C. These results strongly support **H1**.

To evaluate **H2**, we compared our SAM framework with the Random and M-SemExp baselines; recall that the latter was obtained by adapting a state of the art deep RL framework for ON [4]. Table II summarizes the corresponding results for 2-ON and 3-ON tasks. We observed that our SAM framework provided substantially better performance in all three measures considered. Also, performance improved when the number of target object classes increased from 2-ON to 3-ON because of the associated increase in the maximum number of timesteps and our framework's attempt to concurrently reduce the distance to an instance of all target object classes. These results strongly support **H2**.

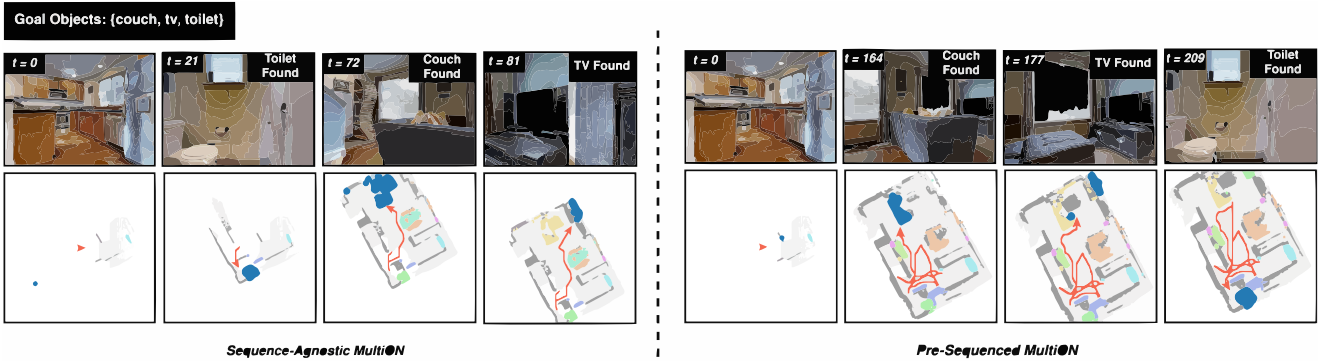

Fig. 3: Qualitative comparison of the performance of our SAM framework (left) with PSM (right) in an episode of the MultiON task with three target object classes (couch, tv, toilet). Our framework results in a smaller number of steps.

| Method | Success (%) ↑ | | Sub-success (%) ↑ | | G-SPL (%) ↑ | |
|---|---|---|---|---|---|---|
| | 2-ON | 3-ON | 2-ON | 3-ON | 2-ON | 3-ON |
| Random | 3.3 | 4.7 | 11.5 | 14.2 | 0 | 0 |
| M-SemExp | 60.5 | 61.7 | 73.1 | 76.6 | 30.5 | 29.8 |
| SAM (ours) | **70.7** | **72.3** | **82.5** | **86.9** | **39.3** | **39.3** |

TABLE II: Our SAM framework provides significantly better performance than the *Random* and *M-SemExp* baselines on three key measures; results averaged over 200 episodes of each of the five testing scenes.

| Timesteps | M-SemExp | | SAM (our method) | |
|---|---|---|---|---|
| | Success (%) | Sub-success (%) | Success (%) | Sub-success (%) |
| 600 | 60.5 | 73.1 | **70.7** | **82.5** |
| 300 | 49.6 | 67.2 | **60.3** | **77.1** |
| 200 | 34.7 | 56.5 | **48.6** | **70.2** |

TABLE III: **2-ON Task:** Comparison of our SAM framework with the M-SemExp baseline for different values of the maximum number of timesteps allowed in each episode.

| Timesteps | M-SemExp | | SAM (our method) | |
|---|---|---|---|---|
| | Success (%) | Sub-success (%) | Success (%) | Sub-success (%) |
| 1000 | 61.71 | 76.6 | **72.3** | **86.9** |
| 500 | 41.34 | 64.6 | **63.7** | **84.3** |
| 300 | 32.49 | 55.0 | **45.3** | **76.4** |

TABLE IV: **3-ON Task:** Comparison of our SAM framework with the M-SemExp baseline for different values of the maximum number of timesteps allowed in each episode.

**Ablation studies:** Next, we performed ablation studies to further explore the effect of the maximum number of allowed timesteps on the performance of our framework compared with the M-SemExp baseline. Specifically, we varied the maximum permissible number of timesteps from 200-600 for the 2-ON task, and from 300-1000 for the 3-ON task, with the results summarized in Table-III and Table-IV respectively. We observed that the degradation in performance as the maximum number of timesteps is reduced was less with our framework than with M-SemExp. These results further reinforced the fact that objects are localized in fewer steps as a result of pursuing a sequence-agnostic approach, and of encouraging the robot to concurrently reduce the distance to an instance of multiple target object classes. Due to the lack of varied objects in the existing data sets (i.e., scenes), for 4-ON, a subset of the data was used and our framework still provided better results than the M-SemExp baseline in each of the different timesteps tested upon. For example, the success rate in 800 timesteps for the M-SemExp baseline and our SAM framework were 61% and 64.6% (respectively) for four target object classes. These results provide further support for hypothesis **H2**.

## VI. DISCUSSION AND FUTURE WORK

We described a deep reinforcement learning-based framework for Sequence-Agnostic Multi-object Navigation (SAM). Object Navigation (ON) methods search for an instance of one object class, and state of the art methods for Multi-Object Navigation (MultiON) are pre-sequenced (PSM), i.e., they are given the order in which the target object classes are to be explored. Our SAM framework extends our prior deep (actor-critic) reinforcement learning framework for ON [5], and includes a suitable reward specification that encourages the desired sequence-agnostic operation. It enables the robot to build on prior experiences in environments similar to the target environment, and to concurrently (and greedily) reduce the distance to an instance of each of the target object classes. Experimental evaluation using realistic scenes from the Gibson dataset in Habitat 3D simulation environment demonstrated a substantial improvement in performance compared with PSM, a baseline that selected actions randomly, and an extension of the SemExp method for ON [4] to MultiON. Future work will explore more complex environments and additional target object classes. Furthermore, we will investigate the use of such a framework on a physical robot assisting humans in practical applications such as Telepresence [27].